\documentclass{article}

\PassOptionsToPackage{numbers, compress}{natbib}

\usepackage[final]{neurips_2023}

\usepackage[utf8]{inputenc} 
\usepackage[T1]{fontenc}    
\usepackage{hyperref}       
\usepackage{url}            
\usepackage{booktabs}       
\usepackage{amsfonts}       
\usepackage{nicefrac}       
\usepackage{microtype}      
\usepackage{xcolor}         
\usepackage{amssymb}
\usepackage{tablefootnote}
\usepackage{lscape}         
\usepackage{afterpage}      
\usepackage{multirow}
\usepackage{graphicx}

\title{LLM-jp: \\ A Cross-organizational Project for the Research and Development of Fully Open Japanese LLMs}

\author{%
  LLM-jp\thanks{Please cite this paper as ``LLM-jp (2024)''. Contribution statements can be found at the end of the document. Correspondence regarding this paper can be sent to \texttt{llm-jp@nii.ac.jp}.} \\
}

\begin{document}

\maketitle

\begin{abstract}
This paper introduces LLM-jp, a cross-organizational project for the research and development of Japanese large language models (LLMs).
LLM-jp aims to develop open-source and strong Japanese LLMs, and as of this writing, more than 1,500 participants from academia and industry are working together for this purpose.
This paper presents the background of the establishment of LLM-jp, summaries of its activities, and technical reports on the LLMs developed by LLM-jp.
For the latest activities, visit \url{https://llm-jp.nii.ac.jp/en/}.
\end{abstract}

\section{Introduction}

Large language models (LLMs), exemplified by GPT-4~\cite{openai2024gpt4}, demonstrate remarkable capabilities.
LLMs have achieved many long-standing goals of traditional natural language processing (NLP), shifting the primary focus of NLP research towards elucidating their intelligence, ensuring their safety, and exploring their integration and coexistence with humans in society.

However, there exist significant issues with LLMs.
First, the research and development of LLMs require significant computational resources and substantial budgets, predominantly controlled by a few major organizations.
Moreover, the specifics of the strongest models --- including their architecture, pre-training corpus, training methodologies, and tuning data --- are no longer publicly accessible.
Additionally, several critical issues, such as hallucination and safety, must be addressed for LLMs to achieve widespread societal acceptance in the future.

There are also national concerns specific to Japan.
The representation of Japanese in the GPT-3 dataset is just 0.11\%\footnote{\url{https://github.com/openai/gpt-3/blob/master/dataset_statistics/languages_by_word_count.csv}}, which results in inferior comprehension and generation of Japanese compared to English.
Furthermore, there is a worry that Japanese culture and activities may be overshadowed if models predominantly trained in English become the global standard.
From an economic security perspective, it is crucial to consider the potential outflow of Japan's intellectual assets when entirely relying on foreign models.

Against this background, LLM-jp started in May 2023 with the objective of developing Japanese LLMs on our own.
The research and development of LLMs is now a big science in terms of both computational and human resources.
Recognizing the need for widespread collaboration, we opted for complete transparency and decided to make everything openly available, from our models, corpora, and fine-tuning data to our discussions and failures, for both non-commercial and commercial use.

LLM-jp began as a small study group of about 30 NLP researchers.
LLM-jp garnered increasing support for its concept over time, growing to over 1,500 participants by June 2024.
Study groups have been held monthly since the establishment of LLM-jp in a hybrid (in-person and online) manner, to introduce the latest advances in LLMs and present the activity reports from LLM-jp.

For the development of LLMs, three working groups (WGs) were first established: the Corpus Building WG, Model Building WG, and Fine-tuning and Evaluation WG.
Subsequently, the Computational Infrastructure WG was formed to address computational infrastructure challenges.
Weekly online meetings and Slack discussions facilitated communication among the groups.
As the project evolved, the Academic Domain WG and Safety WG were also created.

Our first model suite, which we call the LLM-jp model suite v1.0, was released on October 20th, 2023.
Subsequently, we released the next model suite, called the LLM-jp model suite v2.0, on April 30th, 2024.
Each model suite provides an LLM with 13B parameters along with its fine-tuned variants.
We have made them public with their pre-training corpora and fine-tuning datasets.

In the following, we present the activities of the main WGs that played a central role in the construction of our LLMs and future prospects.

\begin{figure}
    \centering
    \includegraphics[width=0.75\columnwidth]{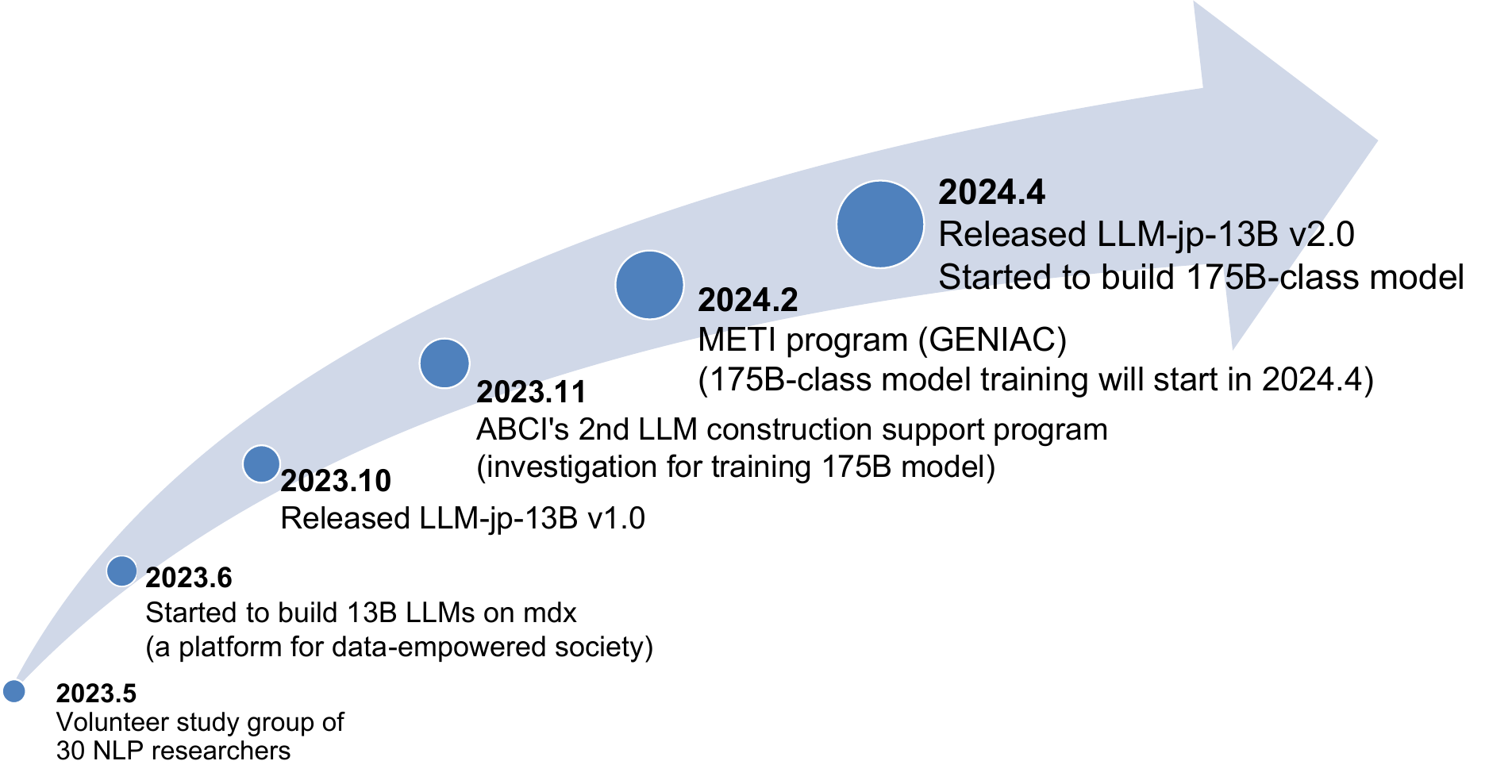}
    \caption{Timeline of key activities in LLM-jp.}
    \label{fig:llm-jp-overview}
\end{figure}

\section{Corpus Building WG}

\subsection{Overview}

The main role of the Corpus Building WG is to build a pre-training corpus and a tokenizer needed for LLM construction and pass them to the Model Building WG.

In the following subsections, we describe our work for the pre-trained models in our model suites v1.0 and v2.0.
Then, we explain the corpus search function, which is one of our advantages. Finally, we summarize our ongoing and future work.

\subsection{Work for Pre-trained Model v1.0}
\label{sec:corpus-v1.0}
Our initial milestone was to develop the model suite v1.0, and the Corpus Building WG worked on preparing a pre-training corpus to train the pre-trained model v1.0, the LLM with 13B parameters within this suite.
The main purpose of this development was to experience the entire development process of an LLM as soon as possible.

To this end, we decided to use a mixture of readily available Japanese, English, and code corpora as our pre-training corpus.
As for the corpus size, we followed the Chinchilla scaling law~\cite{hoffmann2022training}, which suggests using roughly 20 tokens per parameter.
Eventually, we constructed the corpus v1 consisting of over 260B tokens. The statistics of this corpus are listed in Table~\ref{table:corpusv1}. From this corpus, we extracted a pre-training dataset that consists of 130B Japanese, 130B English, and 10B code tokens, resulting in a total of 270B tokens.

\begin{table}[t]
\centering
\caption{Statistics of sub-corpora in the corpus v1.}
\label{table:corpusv1}
\begin{tabular}{l|lr}
\toprule
Language & Sub-corpus & Tokens \\
\midrule
\multirow{2}{*}{Japanese} & Wikipedia & 1B \\
 & mC4 & 136B \\
\midrule
\multirow{2}{*}{English} & Wikipedia & 5B \\
 & Pile & 176B \\
\midrule
Code & Stack & 148B \\
\bottomrule
\end{tabular}
\end{table}

\begin{table}[t]
\centering
\caption{Filters and conversions used for the corpus v1.}
\label{table:corpusv1filter}
\begin{tabular}{l|l}
\toprule
Filter / Conversion & Description  \\
\midrule
HasValidUrlDomain & Filter out documents with URLs from domains rarely used in Japan. \\
IsNotJapanese & Filter out documents that do not contain hiragana or katakana characters. \\
IsNotEthical & Filter out documents that include toxic and/or offensive words. \\
RemoveUrl & Remove URLs from documents. \\
RemoveCode & Remove code-like text spans from documents. \\
\bottomrule
\end{tabular}
\end{table}

As for the Japanese portion, we used the Japanese parts of Wikipedia and the multilingual C4 (mC4) dataset~\cite{xue-etal-2021-mt5}.
Since the Japanese part of mC4 was noisy, we filtered out documents that were considered low-quality or harmful.
Table~\ref{table:corpusv1filter} shows filters adopted for this purpose.
For the English and code portions, we utilized the Pile dataset~\cite{gao2020pile} and the Stack dataset~\cite{kocetkov2022stack}, respectively.
To adjust the corpus size, we sampled documents from these two sources accordingly.

We developed tokenizers based on SentencePiece with the unigram mode~\cite{kudo2018subword}.
As a multi-lingual tokenizer considering Japanese, we first explored the tokenizer developed in the project ``Development of a distributed parallel learning method for large-scale language models in the policy-oriented framework of the supercomputer Fugaku''\footnote{\url{https://www.titech.ac.jp/english/news/2023/066798}}, which we refer to as the tokenizer v1.0.
The construction process is as follows:
\begin{enumerate}
    \item Preparing training data to construct the tokenization models for each language (i.e., Japanese and English).
    \item In order to prevent the tokenization models from learning tokens longer than Japanese word boundaries, Japanese data was pre-tokenized using the morphological analyzer MeCab\footnote{\url{https://taku910.github.io/mecab/}} with the Japanese morphological dictionary JumanDIC\footnote{\url{https://hayashibe.jp/tr/mecab/dictionary/juman}}.
    This pre-tokenization specifically aimed to avoid learning tokens such as browser operation phrases, which are frequently included in web corpus, and meaningless long phrases, which are typically used only on specific websites.
    Pre-tokenization was also performed for sequences including characters other than the alphabet, hiragana, katakana, and kanji into a sequence of single characters to prevent the constructed vocabulary from including tokens with a sequence of symbols and numbers.
    \item Constructing SentencePiece models of the unigram tokenizer for Japanese and English using the pre-processed training data, independently. 
    \item Merging the vocabularies of the above two tokenization models, removing duplicate tokens. 
    \item Re-estimating unigram scores of tokens in the merged vocabulary with the EM algorithm over the training data\footnote{Existing implementation of multigram language model~\cite{deligne1995language} was used, which is available at \url{https://github.com/tatHi/multigram}.}. Here, data without pre-processing was used to enable the final tokenization model to be used without any pre-tokenization.
\end{enumerate}
Although the construction process seems complicated, the obtained model can be used as a pure SentencePiece model.
This multi-step process for the model construction enables us to control the ratio of the vocabulary size for each language.

However, because the tokenizer v1.0 was originally developed for the Fugaku project, we needed to re-train the tokenizer model with the corpus used in the LLM-jp project, the corpus v1.
In addition, some specifications of the tokenizer v1.0, such as the handling of white spaces and line breaks, were left open for discussion.

Therefore, based on the idea of the tokenizer v1.0, we constructed the tokenizer v2.1 for use in the model suite v1.0 by using a subset of the corpus v1 and extending the target languages to Japanese, English, and code.
Besides, we adjusted the handling rules of white spaces, line breaks, and special tokens, which resulted in efficient tokenization in the corpus v1.
The vocabulary of the tokenizer v2.1 is constructed from 30,000 tokens for Japanese, 20,000 tokens for English, and 10,000 tokens for code.
The final size of the vocabulary is approximately 50,000, which indicates that about 10,000 tokens are duplicated among the three vocabularies.

The corpus v1 and tokenizer v2.1 were handed over to the Model Building WG in August 2023 and used for pre-training.
The Model Building WG requested the highest quality corpus used at the end of the pre-training.
In response, we applied the filtering methods described earlier with stricter thresholds to the original corpora and extracted 27B high-quality tokens.

The corpus v1 is publicly available\footnote{\url{https://gitlab.llm-jp.nii.ac.jp/datasets/llm-jp-corpus-v1}}.
The code to construct the corpus is also released to the public\footnote{\url{https://github.com/llm-jp/llm-jp-corpus}}.
Besides, the tokenizer v2.1 and its corresponding scripts are available for download\footnote{\url{https://github.com/llm-jp/llm-jp-tokenizer/releases/tag/v2.1}}.

\subsection{Work for Pre-trained Model v2.0}
\label{sec:corpus-v2.0}

\begin{figure}[t]
\centering
\includegraphics[width=\textwidth]{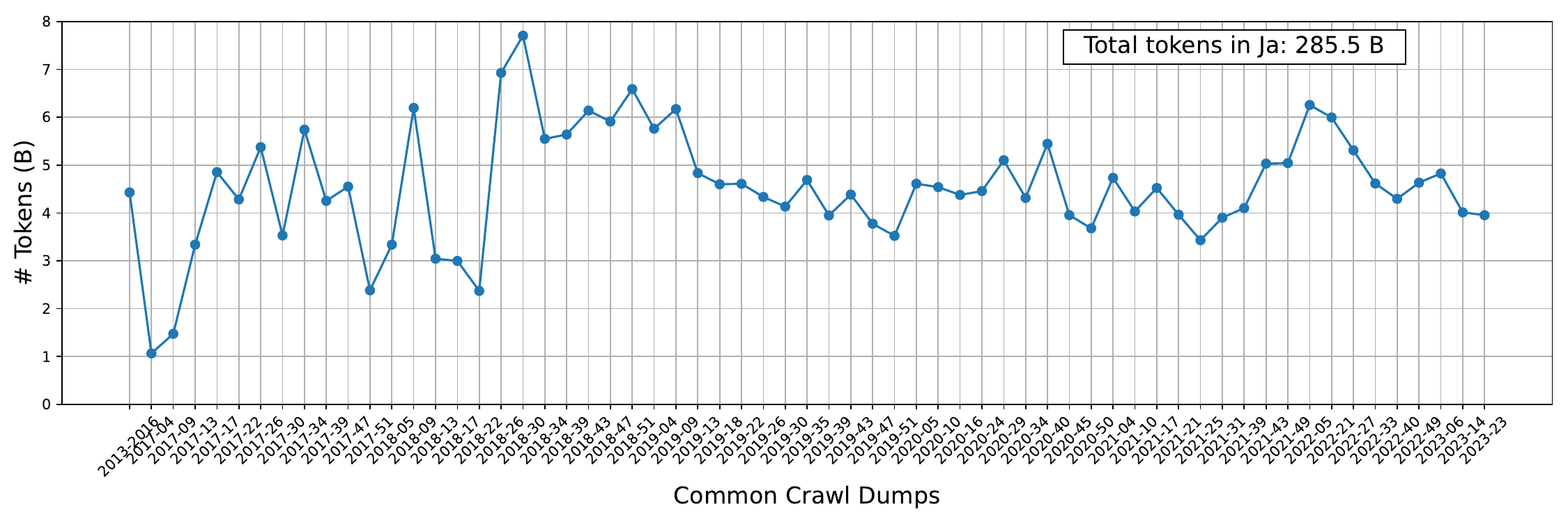}
\caption{Token counts over Common Crawl dumps in the v2 Japanese corpus.}
\label{fig:tokencountsv2}
\end{figure}

\begin{table}[t]
\centering
\caption{Filters and conversions used in Uzushio for the corpus v2.}
\label{table:corpusv2filter}
\begin{tabular}{l|l}
\toprule
Filter / Conversion & Description  \\
\midrule
DocLength & The length of each document.  \\
HiraganaRatio & The upper and lower limits filter of the appearance rate of\\
&Hiragana characters. \\
LinkCharRatio & The upper and lower limits filter of the appearance rate of\\
&hyperlinks in characters. \\
MergeListTag        & Summarizing HTML lists into one paragraph. \\
MarkdownizeHeading  & Converting HTML headings into the Markdown format. \\
NoContentDOM        & Filtering HTMLs with navigational DOM. \\
LargeFreqParagraphs & Removing frequent paragraphs in documents. \\
KenLMParagraphPerplexity & Perplexity-based filter, tokenization by Sudachi\footnote{\url{https://github.com/WorksApplications/Sudachi}}. \\
CompressionRate & The upper and lower limits filter of the zip-compression rate. \\
WordTypes       & Document filter by inappropriate word lists. \\
DocLength       & Document length in characters. \\
DeduplicateDocumentsPercentile  & De-duplication with probabilistic document  identification \\
&by SimHash. \\
\bottomrule
\end{tabular}
\end{table}

To develop the LLM with 13B parameters included in our model suite v2.0, called the pre-trained model v2.0, we created a larger and higher-quality corpus, termed the corpus v2.

To construct a Japanese corpus to this end, we extracted Japanese documents from the entire Common Crawl and applied deduplication and filtering for them.
The corpus v2 construction script was developed in Uzushio\footnote{\url{https://github.com/WorksApplications/uzushio}}, an Apache Spark-based corpus preprocessing tool developed for processing billion-token scale training corpus from web data such as Common Crawl.
Uzushio provides a framework for processing such as similarity-based duplicate detection and filtering.
Table~\ref{table:corpusv2filter} summarizes the filters and conversions performed to construct the Japanese portion of the corpus v2.
The filtering pipeline consisted of deduplication and rule-based filtering steps.
In de-duplication, Uzushio performs similarity-based document identification based on the SimHash algorithm.
This allows Uzushio to apply multiple strengths of de-duplication to documents from a web corpus.
The statistics of the Japanese corpus from Common Crawl dumps are presented in Figure~\ref{fig:tokencountsv2}.
We used the publicly available Common Crawl dumps from 2013 to the middle of 2023.
We merged the Common Crawl dumps from 2013 to 2016 because they included fewer Japanese documents than the later dumps.
The total extracted Japanese tokens were 285.5B\footnote{In the training of the pre-trained model v2.0, we sampled approximately 130B tokens, following the Chichilla scaling law.}.
Further analyses on the v2 corpus are discussed in \citet{enomoto-etal-2024-investigating}.

As for the English and code portions, we used the Pile and Stack datasets, respectively, following the corpus v1.
Besides, we included Japanese and Wikipedia as high-quality text corpora in the corpus v2.

The corpus v2 has been made publicly available.\footnote{\url{https://gitlab.llm-jp.nii.ac.jp/datasets/llm-jp-corpus-v2}}

As for the tokenizer, we newly developed the tokenizer v2.2.
The training flow of the tokenization model is the same as that of the tokenizer v2.1.
The size of the vocabulary was expanded to 96,867\footnote{You can see that the vocabulary size is 97,024 after loading the language model (i.e., vocab\_size in config.json). This is a result of rounding up the vocabulary size to multiples of 256 to make the SoftMax layer training process on the GPU efficient.}.
Besides, while the tokenizer v2.1 used a single token per character for symbols to conserve vocabulary, which resulted in over-segmentation of English and code text and reduced tokenization efficiency, in the tokenizer v2.2, the vocabulary is constructed in a way that allows for symbol sequences, and the tokenization efficiency is improved, especially for English and code text.

\subsection{Corpus Search}

In addition to corpus construction, the Corpus Building WG is also working on developing a corpus search function, aiming to attribute generated text to the training corpus.
This function will be used to analyze generated texts and potentially uncover the principles of LLMs from the perspective of the training corpus.
For example, we plan to use this system to investigate the causes of hallucinations in generated text.

Currently, two search algorithms are being explored: sparse vector search and dense vector search.
Sparse vector search retrieves documents based on the superficial similarity between texts.
It is particularly effective when the generated texts contain distinctive words.
Additionally, it also helps identify verbatim memorization~\cite{carlini2020extracting} in generated texts.
Dense vector search, on the other hand, retrieves documents based on the similarity between text embeddings computed by pre-trained text embedding models.
Compared to sparse vector search, dense vector search excels at considering the meaning of texts.
Furthermore, by using multilingual text embedding models (e.g., LaBSE~\cite{feng-etal-2022-language}), it can retrieve semantically similar documents across different languages, which helps analyze the cross-lingual transfer ability of LLMs~\cite{philippy-etal-2023-towards}.

\subsection{Ongoing and Future Work}
\label{sec:Corpus:Ongoing and Future Work}
We decided to build a 175B-class model as the next target of model building in LLM-jp, and are now building the corpus v3. This new corpus will consist of approximately 2T tokens that cover Japanese, English, some Asian languages, and code.

In our corpora, the mixing ratio of Japanese and English is set at 50-50, but we believe that further study is needed on the mixing ratio and the size of the corpora. In addition to Wikipedia and web documents, we are negotiating with relevant organizations to use high-quality corpora and corpora from various domains, such as scientific and technical papers, patent documents, and domain documents from the medical field.

\section{Computational Infrastructure WG}
\label{sec:infrastructure}

LLM-jp used \texttt{mdx}\footnote{\url{https://mdx.jp/en/}} as the computing resource for training LLMs. \texttt{mdx} is a cloud computing environment consisting of CPUs and GPUs leveraging virtualization technologies~\cite{suzumura2022cbdcom}. \texttt{mdx} provides users with isolated tenants involving virtual machines, virtual networks, and storage. \texttt{mdx} is operated by 11 national organizations in Japan, including nine national universities, the National Institute of Informatics, and the National Institute of Advanced Industrial Science and Technology. In May 2023, \texttt{mdx} had just started official operation and had GPU resources available; thus, we decided to use \texttt{mdx} to build the LLM-jp model.

A GPU node on \texttt{mdx} has eight NVIDIA A100 40GB SXM model GPUs and two Intel Xeon Platinum 8369 model CPUs.
The network is a full-bisection spine-leaf topology where nodes are connected with four 100 Gbps links.
The network supports RoCE (RDMA over Converged Ethernet), an Ethernet-based RDMA protocol, over Virtual eXtensible LAN (VXLAN) for network virtualization.
Thus, GPUs can use RDMA to communicate with other GPUs.
In the LLM-jp configuration, we built a GPU cluster with 16 nodes (128 GPUs) and allocated all GPUs and two 100 Gbps NICs to each virtual machine.

We faced performance issues when we constructed the cluster with 128 GPUs. When we built the pre-trained model v1, there were packet losses in the GPU data communication because ECMP (Equal Cost Multi Path) was not working properly for RoCE packets on the network switch. The performance issue could not be resolved by the start date of the pre-training of the pre-trained model v1, so we reduced the scale of the cluster from 16 nodes (128 GPUs) to 12 nodes (96 GPUs). For the pre-trained model v2.0, we fixed the ECMP issue and used all 16 nodes. Computational Infrastructure WG will share the operational expertise on GPU clusters with other projects.

\section{Model Building WG}

\subsection{Overview}
The role of the Model Building WG is to pre-train language models.
The main tasks include:
\begin{enumerate}
    \item preprocessing the pre-training corpus (such as converting it into a binary format for faster data loading during pre-training),
    \item performing the pre-training, and
    \item converting the checkpoints from the pre-training into a model format that is suitable for fine-tuning.
\end{enumerate}  

The following subsections describe how we built the pre-trained models v1.0 and v2.0.
Table~\ref{table:pre-training-configuration} summarizes the configuration for these models.

\begin{table}[t]
\centering
\caption{Configurations for the pre-trained models v1.0 and v2.0}
\label{table:pre-training-configuration}
\begin{tabular}{lrr}
\toprule
\ &v1.0 &v2.0\\
\midrule
Model Size&13B params & 13B params\\
Corpus size&270(+27)B & 255B \\
Corpus version & v1 & v2\\
Computational environment & \texttt{mdx} 12 nodes & \texttt{mdx} 16 nodes\\
Pre-training tool & Megatron-DeepSpeed & Megatron-LM \\
Base model architecture & GPT2 & Llama2\\
Tokenizer version& v2.1 & v2.2\\
Vocabulary size & 50k & 100k\\
\bottomrule
\end{tabular}
\end{table}

\subsection{Work for Pre-trained Model v1.0}
\label{sec:model-v1.0}
In May 2023, when this project started, the Model Building WG began its activities with the aim of building and releasing a 13B-parameter model specifically focusing on Japanese by autumn or winter 2023.
At the start of activities in May 2023, no one in the Model Building WG had solid knowledge or experience in pre-training language models with over 10B parameters using more than 10 computing nodes. 
Therefore, all participants in the WG experienced what was necessary for pre-training step by step, gaining knowledge and experience through a process of trial and error.

First, to pre-train a language model, we need to prepare a training program (code).
While there was the option to develop our own training program, our goal was to build a 13B-parameter model within a few months, making the option infeasible.
At the start of the project in May 2023, there were several tools available for pre-training language models with over 10B parameters, so we decided to use them.
Specifically, we considered Megatron-DeepSpeed\footnote{\url{https://github.com/microsoft/Megatron-DeepSpeed}}, GPT-NeoX\footnote{\url{https://github.com/EleutherAI/gpt-neox}}, and LLM Foundry\footnote{\url{https://github.com/mosaicml/llm-foundry}} as candidates.
Pre-training requires massive computing resources, such as using more than 10 GPU nodes for over 10 days.
Therefore, there was not enough time to run multiple training sessions simultaneously or to use multiple tools to create and compare several models side-by-side.
Considering several factors, including the fact that several participants had experience with it and that developers involved in DeepSpeed had been participating in LLM-jp activities from the beginning, we regarded Megatron-DeepSpeed as the primary tool for Model Building WG's activities.
However, it was also decided to use GPT-NeoX and LLM Foundry in parallel up to the verification of training speed and stability.
Each tool was assigned to a person in charge, and teams were formed to compare the results. Eventually, we chose to build the pre-training model using Megatron-DeepSpeed, as we did not find GPT-NeoX and LLM Foundry to be clearly superior in terms of execution speed or training stability (just to clarify, GPT-NeoX and LLM Foundry were not inferior).

First of all, the training speed of the language model was verified.
For example, when training a language model with more than 10B parameters on a training corpus of more than 100B tokens, it is usually necessary to use a computing environment of more than 10 GPU nodes to finish the training in a realistic time.
Even if the same language model learning configuration is used, the learning speed can vary greatly depending on the characteristics of the computer cluster environment.
Therefore, it is necessary to find the appropriate learning settings for each computer environment used.
The details of the computing environment and the various issues related to it are summarised in Section~\ref{sec:infrastructure}.
We searched for the optimal setting according to FLOPS (floating point operations per second), an index that is independent of the size of the model and differences in the computer environment and thus often used in existing research as a measure of learning speed.
In Megatron-Deepspeed, there are many configurable settings related to the learning speed of language models, including model parallelism (tensor parallel, pipeline parallel)~\cite{lepikhin2021gshard,DBLP:journals/corr/abs-1909-08053}, data parallelism, batch size, and a setting called ZeRO~\cite{DBLP:conf/sc/RajbhandariRRH20}, which mainly determines the trade-off between GPU/CPU memory utilization and speed.
Various settings were prepared by combining the values of each item, and measured learning speeds were collected for each.
Finally, the setting that produced the most stable and highest FLOPS value was adopted.

By measuring the actual processing speed, we predicted the total time required for the model construction once the size of the training corpus is determined.
The total learning time at that time was predicted as follows: 
Using a 12 node \texttt{mdx} computing cluster to train a model with 13B parameters, the measured processing speed was 170K tokens/second on average. 
Therefore, the estimated total time required for training the 13B parameter model with a corpus of 270B tokens was roughly 441.2 hours or about 18.4 days.

We had been preparing to learn a language model using a training corpus of 270B tokens, but as the volume of the training corpus was expected to increase, we considered a learning method that would enable continuous pre-training even if the training corpus increased sequentially. 
Here, we tried a method in which the training data of 270B tokens is divided roughly into 10 chunks of 27B tokens, and these tokens are trained one by one. 
Assuming that about one trillion tokens of data would be learned in the future, we applied a setting for learning one trillion tokens to the learning rate scheduler, which is the cosine decay scheduler typically used in the literature of pre-training language models.
We also asked the Corpus Building WG to prepare a training corpus of 27B tokens selected from 270B tokens, which were considered to be of high quality, and used this 27B token training corpus at the end of the pre-training. 
When we trained the final 270 billion tokens, we also rapidly decreased the learning rate to the predefined final learning rate for overall pre-training. 
This decrease started from the learning rate at the end of the preceding 27B tokens training, using the same cosine scheduler but with a different hyperparameter setting.

Another aspect to consider when pre-training language models is the stability of the training. 
In LLM pre-training, we often observe that the model cannot be learned effectively due to loss divergence, often called loss explosion and loss spike. 
At that moment, the mechanism of loss divergence had not been fully elucidated. 
Therefore, we need to explore and use a setting in which loss divergence occurs as little as possible. 
We are basically required to deal with this problem through trial and error, but fortunately, no unresolvable loss divergence occurred in our pre-training.

The pre-trained model v1.0 uses a model architecture based on GPT-2~\cite{Radford-gpt2}. 
Although GPT-2 is a relatively old model architecture, and while a newer one was possible, we deemed it more appropriate to use a well-established and stable one, considering the need for a reliable model for many users. 
Additionally, converting the model checkpoints to a format compatible with the Hugging Face Transformers library\footnote{\url{https://github.com/huggingface/transformers}} is a common practice, making it crucial to ensure the model can be converted. 
Unfortunately, the Transformers library does not support the Megatron-DeepSpeed model format used in our training, so a conversion script is needed. 
From this perspective, while Megatron-DeepSpeed offers a script for converting to the Hugging Face Transformers format, it only supports GPT-2-based models. 
Therefore, without a custom conversion script, we could only use GPT-2-based models with Megatron-DeepSpeed. 
Given our limited resources and the fact that this was LLM-jp's first attempt, we concluded that using the GPT-2 model was the safest choice.

Following the various studies described above, the preliminary learning of the language model for public use began in earnest around the end of August. In practice, learning the target 13B parameter model out of the blue was also risky, so learning the 1.3B parameter model was carried out as a pre-production exercise.
Eventually, the pre-training of the 13B parameter model took 26 days.
During the training process, there was trouble that the training stopped several times, and it was necessary to restart the training manually. If the training had proceeded without any problems, it could have been carried out in about 21 days at the shortest. The model was then handed over to the Fine-tuning and Evaluation WG, which completed the work of building the model v1.0 for publication in the Model Building WG.

\subsection{Work for Pre-trained Model v2.0}
As mentioned in the previous section, the pre-trained model v1.0 was our initial attempt, and we had a time constraint for its construction and release. 
This means that our primary focus was on quickly building the model on schedule rather than investigating how to obtain a world-class, high-quality model. 
To identify a better pre-training configuration for the pre-trained model v2.0, we conducted experiments prior to beginning its construction.

\subsubsection{Preliminary Experiments: Towards Better Pre-trained Model v2.0}

We have changed several pre-training configurations of the pre-trained model v1.0 for model v2.0 since we aimed to improve the overall performance.
Regarding the model architecture, we decided to replace GPT-2 used in model v1.0 with the Llama architecture, which was starting to gain wide adoption at that time.
We conducted experiments to determine the best configuration. 
The primary factors of evaluation included the vocabulary size and pre-training corpus type.
For vocabulary size, we compared 50k and 100k while the tokenizer was given and fixed to v2.2.
As for the pre-training corpus type, we examined three types of Japanese sub-corpora: the Japanese part in the corpus v1 used for constructing the model v1.0, the Swallow corpus\footnote{\url{https://tokyotech-llm.github.io/swallow-corpus}} used for continual pre-training from Llama 2, and the corpus v2 prepared specifically for the model v2.0. 
We refer to these three Japanese datasets for pre-training as \texttt{llmjp v1(ja)}, \texttt{Swallow}, and \texttt{llmjp v2$\beta$(ja)}, respectively.
Regarding the English and Code parts of the dataset for the pre-training, we reused the identical sub-corpora to build for the model v1 (Table~\ref{table:corpusv1}).
We sampled approximately 114.5B and 8.7B tokens (under the 100k vocabulary) from these sub-corpora, respectively.

We prepared several configurations based on the comparison factors of vocabulary sizes and pre-training corpus types to clarify the effectiveness of each aspect.
Table~\ref{tab:MCWG_V2experiments} summarizes the configurations used for our preliminary experiments. 
We used Megatron-LM\footnote{https://github.com/NVIDIA/Megatron-LM} for all experiments in this section instead of Megatron-Deepspeed used for building pre-trained model v1.0.

\begin{table}[t]
\centering
\small
\caption{Experimental configurations for comparing the effectiveness of selected training corpus and vocabulary size.}
\label{tab:MCWG_V2experiments}
\begin{tabular}{ccccccc}
\toprule
Exp. & Param. & Japanese    & Japanese    & Tokenizer & Vocab. \\ 
ID  & size    & corpus & corpus size & Version   & size \\ 
\midrule
\texttt{Exp(a)} &  7B &  \texttt{llmjp v1(ja)} & (134B) & v2.2 &50k \\ 
\texttt{Exp(b)} &  7B &  \texttt{Swallow}      & (147B) & v2.2 &50k \\ 
\texttt{Exp(c)} &  7B &  \texttt{llmjp v2$\beta$(ja)} & (135B) & v2.2 &50k \\ 
\texttt{Exp(d)} &  7B &  \texttt{llmjp v1(ja)} & (131B) & v2.2 & 100k \\ 
\texttt{Exp(e)} & 13B &  \texttt{llmjp v2$\beta$(ja)} & (135B) & v2.2 &50k \\ 
\texttt{Exp(f)} &     7B & \texttt{llmjp v2$\beta$(ja)} & (250B) & v2.2 &50k \\ 
\texttt{Exp(g)} & 13B & \texttt{llmjp v2$\beta$(ja)} & (131B) & v2.2 & 100k \\ 
\bottomrule
\end{tabular}
\end{table}
\begin{table}[t]
\centering
\small
\tabcolsep 2pt
\caption{Experimental results for comparing the effectiveness of selected training corpus and vocabulary size. 
In the title raw, llm-jp and JVQA represent the \texttt{llm-jp-eval} benchmark and the Japanese Vicuna QA benchmark, respectively.}
\label{tab:MCWG_V2exp_results}

\begin{tabular}{cccc}
\texttt{(C1)} Corpus type & \texttt{(C2)} Vocab. size & \texttt{(C3)} Model size & \texttt{(C4)} Corpus size
\\
\begin{tabular}[t]{ccccccc}
\toprule
Exp. ID & llm-jp & JVQA \\ 
\midrule
\texttt{Exp(a)} & 0.539 & 40.36\\
\texttt{Exp(b)} & 0.561 & 35.38\\
\texttt{Exp(c)} &\bf 0.562 &\bf 43.52\\
\bottomrule
\end{tabular}
&
\begin{tabular}[t]{ccccccc}
\toprule
Exp. ID & llm-jp & JVQA \\ 
\midrule
\texttt{Exp(c)} &\bf 0.562 &\bf 43.52\\
\texttt{Exp(d)} &    0.548 &    34.26\\
\midrule
\texttt{Exp(e)} &\bf 0.577 &    47.00\\
\texttt{Exp(g)} &    0.576 &\bf 50.74\\
\bottomrule
\end{tabular}
&
\begin{tabular}[t]{ccccccc}
\toprule
Exp. ID & llm-jp & JVQA \\ 
\midrule
\texttt{Exp(c)} &    0.562 &    43.52\\
\texttt{Exp(e)} &\bf 0.577 &\bf 47.00\\
\bottomrule
\end{tabular}
&
\begin{tabular}[t]{ccccccc}
\toprule
Exp. ID & llm-jp & JVQA \\ 
\midrule
\texttt{Exp(c)} &\bf 0.562 &    43.52\\
\texttt{Exp(f)} &    0.556 &\bf 49.88\\
\bottomrule
\end{tabular}

\end{tabular}
\end{table}

The following four perspectives of comparison (\texttt{(C1)}, \texttt{(C2)}, \texttt{(C3)}, and \texttt{(C4)}) are the primary intentions of our preliminary experiments:
\begin{itemize}
\item[\texttt{(C1)}] Comparing \texttt{Exp(a)}, \texttt{Exp(b)}, and \texttt{Exp(c)}, we attempted to investigate which one of the Japanese corpora can be better in terms of pre-training.
Remember that the corpus v1 (ja), \texttt{Swallow}, and \texttt{llmjp v2$\beta$(ja)} can contain identical and near identical texts.
Therefore, it's not as straightforward as simply combining these three corpora into one for pre-training purposes. 
This is because changes in data distribution and the inclusion of duplicate data could potentially harm and degrade the pre-training process.

\item[\texttt{(C2)}] Comparing \texttt{Exp(a)} and \texttt{Exp(d)} and also \texttt{Exp(e)} and \texttt{Exp(g)}, we can see the effectiveness of increasing vocabulary size from 50k to 100k.

\item[\texttt{(C3)}] Comparing \texttt{Exp(c)} and \texttt{Exp(e)}, we can see the effectiveness of increasing model parameter size.

\item[\texttt{(C4)}] Comparing \texttt{Exp(c)} and \texttt{Exp(f)}, we can see the effectiveness of increasing corpus size.
\end{itemize}

After pre-training for each configuration, we performed simple fine-tuning on each pre-trained model and evaluated the performance by llm-jp-eval and Japanese Vicuna QA benchmarks, as introduced in Section~\ref{sec:evaluation}.
Table~\ref{tab:MCWG_V2exp_results} shows the results.
The findings from these results are as follows:
\begin{enumerate}
    \item According to the \texttt{(C1)} result, the corpus v2 (\texttt{llmjp v2$\beta$(ja)}) seems to perform better than the corpus v1 (\texttt{llmjp v1(ja)}) and Swallow corpus.
    \item According to the \texttt{(C2)} result, the performance difference between vocabulary sizes of 50k and 100k seems marginal, and we are unable to determine which is better clearly. 
    \item From the \texttt{(C3)} result, the model size significantly affects the performance; this is the consistent result of common knowledge like scaling laws.
    \item From the \texttt{(C4)} result, the corpus size for pre-training also affects the performance.
\end{enumerate}
These results led to the decision on the model setting for v2.0, described in Table~\ref{table:pre-training-configuration}.

\subsubsection{Constructing Pre-trained Model v2.0}
\label{sec:model-v2.0}
As demonstrated in the preliminary experiment, \texttt{Exp(g)} appears to deliver the best performance.
Therefore, we decided to adopt the model trained in \texttt{Exp(g)} as the pre-trained model v2.0.
Furthermore, with the model trained in \texttt{Exp(g)} being adopted as the pre-trained model v2.0, the training data used in Exp(g) was also finalized as corpus v2.

\subsection{Ongoing and Future Work}
As described in Section~\ref{sec:Corpus:Ongoing and Future Work}, we plan to build a 175B-parameter-class model as the next target of model building in LLM-jp.
In practice, we have already tried pre-study using a GPT-3 compliant model on a trial basis using the LLM construction support program at ABCI\footnote{\url{https://abci.ai/ja/link/llm_support_program2023.html}} and have identified some issues to consider, such as loss-spike. 
We are preparing the implementation to mitigate such issues.
The Model Building WG is diligently working to build a 175B-parameter-class language model, trained with a dataset of over 1T tokens (called the corpus v3), publicly available this autumn.
For this purpose, we have submitted (and been selected) to an LLM construction support program at the Ministry of Economy, Trade and Industry (METI) in Japan, called GENIAC\footnote{\url{https://www.meti.go.jp/policy/mono_info_service/geniac/index.html}}.


\section{Fine-tuning and Evaluation WG}

\subsection{Overview}

This section introduces our efforts on fine-tuning and evaluation of LLMs.
Pre-trained language models can produce natural and fluent text following input text (prompts), but they do not necessarily produce responses that humans would expect in response to the input.
To develop interactive LLMs like ChatGPT, it is essential for them to have the ability to generate appropriate responses to user input; i.e., they need to be \textit{aligned} with human values~\cite{ouyang2022training}.
Alignment is an essential issue in LLM research and development, and fine-tuning is an indispensable step in achieving this.

Evaluation is another critical issue for the development and deployment of LLMs.
A conventional method for evaluating NLP systems has been to design a specific task, such as question answering and machine translation, and to develop test data for each designed task.
However, this method is insufficient for the evaluation of LLMs because LLMs are used in a variety of downstream tasks.
We therefore develop evaluation frameworks that can assess diverse natural language understanding capabilities of LLMs.


\subsection{Fine-tuning}

\begin{table}[t]
    \centering
    \caption{Datasets for fine-tuning. Dagger ($\dagger$) indicates that the dataset was automatically translated from English.}
    \label{table:tuning-datasets}
    \begin{tabular}{l|r|ccc}
        \toprule
        \ & \# of samples & v1.0 & v1.1 & v2.0 \\
        \midrule
        jaster (JA) & 136,605 & $\checkmark$ & - & - \\
        databricks-dolly-15k (EN) & 15,011 & - & $\checkmark$ & $\checkmark$ \\
        databricks-dolly-15k (JA)$^\dagger$ & 15,011 & $\checkmark$ & $\checkmark$ & $\checkmark$ \\
        oasst1 (EN) & 21,164 & - & $\checkmark$ & $\checkmark$ \\
        oasst1 (JA)$^\dagger$ & 21,164 & $\checkmark$ & $\checkmark$ & $\checkmark$ \\
        hh-rlhf (JA)$^\dagger$ & 12,000 & - & $\checkmark$ & - \\
        oasst2 (EN) & 32,702 & - & - & $\checkmark$ \\
        oasst2 (JA)$^\dagger$ & 32,702 & - & - & $\checkmark$ \\
        ichikara-instruction-003-001 (JA) & 2,903 & - & $\checkmark$ & - \\
        ichikara-instruction-004-001 (JA) & 9,103 & - & - & $\checkmark$ \\
        AnswerCarefully v1.0 (JA) & 945 & - & - & $\checkmark$ \\
    \bottomrule
    \end{tabular}
\end{table}

To date, we have released three versions of our fine-tuned models: v1.0, v1.1, and v2.0. The fine-tuned model v1.0 was released alongside the pre-trained model v1.0.
In the fine-tuned model v1.1, which is based on the same pre-trained model v1.0, we improved the instruction-following ability by refining the instruction-tuning data and adding Direct Preference Optimization (DPO), and released it in February 2024.
The fine-tuned model v2.0, released in April 2024, features the use of pre-trained model v2.0 and incorporates fine-tuning that considers safety aspects. This section outlines the methods for constructing each model.
Table~\ref{table:tuning-datasets} summarizes the datasets used for the fine-tuning of each version.

\subsubsection{Work for Fine-tuned Model v1.0} \label{sssec:v1.0}
For the fine-tuned model v1.0, we constructed three types of Japanese instruction data: jaster, databricks-dolly-15k~\cite{DatabricksBlog2023DollyV2}, and OpenAssistant Conversations Dataset (oasst1)~\cite{köpf2023openassistant}. Jaster is a dataset that utilizes existing datasets from Japanese natural language processing (NLP) tasks. Through the accumulation of research in NLP, training and evaluation data for individual NLP tasks such as natural language inference and question answering have been developed and made available. Jaster was constructed by converting these data into a natural language instruction format and corresponding responses. 
The remaining two instruction datasets are machine-translated from English datasets using DeepL\footnote{\url{https://www.deepl.com/}}. While many instruction datasets are available in English, we selected databricks-dolly-15k and oasst1, as they are widely used and provide suitable licenses for LLM-jp.

Upon the release of the fine-tuned model v1.0, we developed and released \texttt{llm-jp-sft}\footnote{\url{https://github.com/llm-jp/llm-jp-sft}}, an open-source tuning tool designed for supervised fine-tuning. This tool supports not only full-parameter fine-tuning but also LoRA~\cite{hu2022lora}-based fine-tuning.

\subsubsection{Work for Fine-tuned Model v1.1}
After the release of the fine-tuned model v1.0, we worked on improving the instruction-following ability and released the model as the fine-tuned model v1.1.

First, we expanded the instruction dataset used. 
The use of English instruction data in addition to non-English one has been reported to improve model performance in non-English languages~\cite{chen-etal-2024-monolingual}. 
Based on this finding, we decided to add original English datasets of databricks-dolly-15k and oasst1. 
Additionally, we incorporated the Japanese instruction dataset, ichikara-instruction (ver 003-001)~\cite{sekine2023ichikara}.
This dataset, distinct from machine-translated datasets, consists of high-quality instruction data created from scratch in Japanese by human annotators (the term ``ichikara'' means ``from scratch'' in Japanese).

Next, we introduced Direct Preference Optimization (DPO)~\cite{rafailov2023direct}, which is designed to generate responses more preferable to the user. 
DPO has been demonstrated to exhibit performance equal to or greater than Proximal Policy Optimization~\cite{schulman2017ppo}, which is the preference optimization method employed in InstructGPT~\cite{NEURIPS2022_b1efde53}, while also offering superior stability and computational efficiency during training. 
We sampled 12,000 instances from hh-rlhf\footnote{\url{https://huggingface.co/datasets/Anthropic/hh-rlhf}} and made them publicly available as hh-rlhf-ja\footnote{\url{https://huggingface.co/datasets/llm-jp/hh-rlhf-12k-ja}}, which was translated into Japanese using DeepL. 
The training code specific to DPO, \texttt{llm-jp-dpo}, has also been made open-source.\footnote{\url{https://github.com/llm-jp/llm-jp-dpo}}

\subsubsection{Work for Fine-tuned Model v2.0}
Upon the release of the pre-trained model v2.0, we further added instruction data. 
The Open Assistant Conversations Dataset Release 2 (oasst2)\footnote{\url{https://huggingface.co/datasets/OpenAssistant/oasst2}} is an English conversational instruction dataset. 
We utilized both the original English version and a Japanese version translated via DeepL. 
Additionally, we used the new version of ichikara-instruction (004-001). 
Moreover, a new instruction dataset, AnswerCarefully, was introduced for enhanced safety. 
For more details on AnswerCarefully, refer to Section~\ref{sec:answer_carefully}.

\subsection{Evaluation Frameworks} \label{sec:evaluation}
Unlike traditional, task-specific NLP systems, LLMs can generally be used in various applications.
It is, therefore, challenging to develop a specific benchmark to evaluate the entire capability of LLMs.
Because of this problem, many evaluation benchmarks for LLMs have been proposed globally \cite{srivastava2023beyond, zheng20223llm}.
However, the number of evaluation benchmarks, like JGLUE~\cite{kurihara-etal-2022-jglue}, for Japanese LLMs was limited when we started developing LLM-jp models.

We have been developing an evaluation framework to aim for multifaceted evaluation rather than depending on a single benchmark.
A variety of benchmark datasets for conventional NLP tasks for Japanese have been proposed to date.
We have therefore constructed \texttt{llm-jp-eval}\footnote{\url{https://github.com/llm-jp/llm-jp-eval}}, an open-source tool for evaluating Japanese LLMs across these individual tasks.
In the same way as constructing jaster, existing datasets for Japanese NLP tasks are converted into prompt-answer pairs.
When evaluating LLMs, prompts are input, and the text predicted by the target LLM is matched with the answers to measure evaluation scores.
We have continuously updated \texttt{llm-jp-eval} from its first release in October 2023, and now the version of \texttt{llm-jp-eval} is 1.3.0\footnote{As of June 2024.}.
Table~\ref{tab:llm-jp-eval_datasets} shows the list of individual evaluation datasets which \texttt{llm-jp-eval} supports.
Table~\ref{tab:llm-jp-eval_results} shows the result of evaluation for LLM-jp models by \texttt{llm-jp-eval}, and see Table~\ref{tab:llm-jp-models} for the model IDs and details for each LLM-jp model.

For the base models without fine-tuning, v1.0-A/B and v2.0-L, we found that v2.0-L achieved the highest score, as we expected.
We found that the evaluation score of v2.0-L is higher than that of fine-tuned models, v2.0-M/N/O.
Because fine-tuning datasets except jaster are made up of non-routine tasks that require long answers, compared to many tasks in \texttt{llm-jp-eval} requiring relatively short answers.
The evaluation scores of v2.0-M/N/O, fine-tuned variants of v2.0, are higher than v1.0-A/B, indicating LLM-jp v2.0 models are improved from v1.0.

For the fine-tuning method, SFT seems better than LoRA in most cases for LLM-jp models.
Jaster is the training split for a part of \texttt{llm-jp-eval} datasets, and indeed the models fine-tuned with jaster show the best score.
Note that we strictly divided jaster and the evaluation datasets in \texttt{llm-jp-eval} to prevent data leaks.
However, it is evident that fine-tuning with training splits also works like supervised learning in traditional machine learning tasks.
This is the reason why we do not use jatser to fine-tune v2.0 models. 

\begin{table}[t]
    \caption{Datasets which \texttt{llm-jp-eval} supports. Category is an identifier used in \texttt{llm-jp-eval}. Version means which \texttt{llm-jp-eval} version starts to support this dataset.}
    \scriptsize
    \centering
    \begin{tabular}{ccccc}
         \toprule
         Category & Dataset & Task & Metrics & Version
         \\
         \midrule
         EL & chABSA\tablefootnote{\url{https://github.com/chakki-works/chABSA-dataset}} & Entity linking & Set F1 & v1.1.0
         \\
         \hline
         \multirow{5}{*}{FA} & \multirow{5}{*}{Wikipedia Annotated Corpus~\cite{Masatsugu2014}} & Reading prediction & Char. F1 & v1.1.0
         \\
          & & Named entity recognition & Set F1 & v1.1.0
         \\
          & & Dependency parsing & Set F1 & v1.1.0
         \\
          & & Predicate-argument structure analysis & Set F1 & v1.1.0
         \\
          & & Coreference resolution & Set F1 & v1.1.0
         \\
         \hline
         \multirow{2}{*}{HE} & MMLU~\cite{hendrycks2021measuring} & \multirow{2}{*}{Human examination} & Exact Match & v1.3.0
         \\
          & JMMLU\tablefootnote{\url{https://github.com/nlp-waseda/JMMLU}} & & Exact Match & v1.3.0
         \\
         \hline
         \multirow{2}{*}{MT} & ALT Parallel Corpus~\cite{thu-etal-2016-introducing} & \multirow{2}{*}{Machine translation} & Comet & v1.3.0
         \\
          & Wikipedia's Kyoto Articles\tablefootnote{\url{https://alaginrc.nict.go.jp/WikiCorpus/index_E.html}} & & Comet & v1.3.0
         \\
         \hline
         MR & MAWPS~\cite{Kaito2023} & Mathematical reasoning & Exact Match & v1.2.0
         \\
         \hline
         MC & JCommonsenseQA~\cite{kurihara-etal-2022-jglue} &  Multiple choice question answering & Exact Match & v1.0.0
         \\
         \hline
         \multirow{5}{*}{NLI} & Jamp~\cite{sugimoto-etal-2023-jamp} & \multirow{5}{*}{Natural language inference} & Exact Match & v1.0.0
         \\
          & JaNLI~\cite{yanaka-EtAl:2021:blackbox} & & Exact Match & v1.0.0
         \\
          & JNLI~\cite{kurihara-etal-2022-jglue} & & Exact Match & v1.0.0
         \\
          & JSeM~\cite{kawazoe2017inference} & & Exact Match & v1.0.0
         \\
          & JSICK~\cite{yanaka-mineshima-2022-compositional} & & Exact Match & v1.0.0
         \\
         \hline
         \multirow{2}{*}{QA} & JEMHopQA~\cite{ishii-etal-2024-jemhopqa-dataset} & \multirow{2}{*}{Question answering} & Char. F1 & v1.0.0
         \\
          & NIILC\tablefootnote{\url{https://mynlp.is.s.u-tokyo.ac.jp/niilc-qa/}} & & Char. F1 & v1.0.0
         \\
         \hline
         RC & JSQuAD~\cite{kurihara-etal-2022-jglue} & Reading comprehension & Char. F1 & v1.0.0
         \\
         \bottomrule
    \end{tabular}
    \label{tab:llm-jp-eval_datasets}
\end{table}

\begin{table}[t]
    \caption{The LLM-jp models to be evaluated. See Table~\ref{table:tuning-datasets} for the details of the fine-tuning datasets. dolly corresponds to databricks-dolly-15k (EN, JA), oasst to oasst1 and 2 (EN, JA), ichikara to ichikara-instruction-003/004-001 (JA), and AC to AnswerCarefully v1.0 (JA). 16x means using 16x augmented dataset.}
    \centering
    \small
    \begin{tabular}{lccccccccc}
         \toprule
         Model ID & Version & Param. & Tuning & jaster & dolly & oasst & ichikara & HH-RLHF & AC \\
         \midrule
         v1.0-A & 1.0 & 1.3b & None & & & & & & \\
         v1.0-B & 1.0 & 13b  & None & & & & & & \\
         v1.0-C & 1.0 & 13b  & SFT  & $\checkmark$ & & & & & \\
         v1.0-D & 1.0 & 13b  & LoRA & $\checkmark$ & & & & & \\
         v1.0-E & 1.0 & 13b  & SFT  & & $\checkmark$ & $\checkmark$ & & & \\
         v1.0-F & 1.0 & 13b  & SFT  & $\checkmark$ & $\checkmark$ & $\checkmark$ & & & \\
         v1.0-G & 1.0 & 13b  & LoRA & & $\checkmark$ & $\checkmark$ & & & \\
         v1.0-H & 1.0 & 13b  & LoRA & $\checkmark$ & $\checkmark$ & $\checkmark$ & & & \\
         \midrule
         v1.1-I & 1.1 & 13b  & SFT  & & $\checkmark$ & $\checkmark$ & $\checkmark$ & & \\
         v1.1-J & 1.1 & 13b  & LoRA & & $\checkmark$ & $\checkmark$ & $\checkmark$ & & \\
         v1.1-K & 1.1 & 13b  & SFT+DPO  & & $\checkmark$ & $\checkmark$ & $\checkmark$ & $\checkmark$ & \\
         \midrule
         v2.0-L & 2.0 & 13b  & None & & & & & & \\
         v2.0-M & 2.0 & 13b  & SFT  & & $\checkmark$ & $\checkmark$ & $\checkmark$ & & \\
         v2.0-N & 2.0 & 13b  & SFT  & & $\checkmark$ & $\checkmark$ & $\checkmark$ & & $\checkmark$ \\
         v2.0-O & 2.0 & 13b  & SFT  & & $\checkmark$ & $\checkmark$ & $\checkmark$ & & 16x \\
         \bottomrule
    \end{tabular}
    \label{tab:llm-jp-models}
\end{table}

\begin{table}[t]
    \caption{The result of evaluation of LLM-jp models by \texttt{llm-jp-eval} v1.3.0. AVR is the average score across all categories. See Table~\ref{tab:llm-jp-eval_datasets} for the details of evaluation categories.}
    \centering
    \small
    \begin{tabular}{lrrrrrrrrrr}
         \toprule
         Model ID & \multicolumn{1}{c}{AVR} & \multicolumn{1}{c}{EL} & \multicolumn{1}{c}{FA} & \multicolumn{1}{c}{HE} & \multicolumn{1}{c}{MC} & \multicolumn{1}{c}{MR} & \multicolumn{1}{c}{MT} & \multicolumn{1}{c}{NLI} & \multicolumn{1}{c}{QA} & \multicolumn{1}{c}{RC}
         \\
         \midrule
         v1.0-A & 0.269 & 0.105 & 0.067 & 0.260 & 0.203 & 0.020 & 0.597 & 0.309 & 0.303 & 0.557
         \\
         v1.0-B & 0.382 & 0.352 & 0.176 & 0.249 & 0.203 & 0.130 & 0.787 & 0.349 & 0.469 & 0.721
         \\
         v1.0-C & 0.507 & 0.188 & 0.071 & 0.301 & 0.884 & 0.136 & 0.604 & 0.911 & 0.544 & 0.923
         \\
         v1.0-D & 0.491 & 0.169 & 0.052 & 0.316 & 0.874 & 0.140 & 0.482 & 0.920 & 0.540 & 0.923
         \\
         v1.0-E & 0.386 & 0.378 & 0.163 & 0.254 & 0.217 & 0.146 & 0.780 & 0.408 & 0.406 & 0.727
         \\
         v1.0-F & 0.536 & 0.276 & 0.140 & 0.307 & 0.849 & 0.168 & 0.714 & 0.909 & 0.535 & 0.924
         \\
         v1.0-G & 0.378 & 0.389 & 0.138 & 0.247 & 0.223 & 0.104 & 0.737 & 0.401 & 0.421 & 0.739
         \\
         v1.0-H & 0.524 & 0.317 & 0.114 & 0.296 & 0.805 & 0.140 & 0.704 & 0.861 & 0.562 & 0.919
         \\ \midrule
         v1.1-I & 0.365 & 0.367 & 0.155 & 0.237 & 0.221 & 0.042 & 0.759 & 0.435 & 0.361 & 0.708
         \\
         v1.1-J & 0.395 & 0.387 & 0.159 & 0.241 & 0.258 & 0.044 & 0.786 & 0.480 & 0.471 & 0.726
         \\
         v1.1-K & 0.350 & 0.351 & 0.151 & 0.236 & 0.225 & 0.042 & 0.774 & 0.359 & 0.330 & 0.678
         \\ \midrule
         v2.0-L & 0.405 & 0.389 & 0.241 & 0.253 & 0.183 & 0.182 & 0.796 & 0.298 & 0.522 & 0.781
         \\
         v2.0-M & 0.387 & 0.350 & 0.196 & 0.250 & 0.186 & 0.216 & 0.785 & 0.316 & 0.421 & 0.759
         \\
         v2.0-N & 0.383 & 0.355 & 0.192 & 0.246 & 0.193 & 0.208 & 0.782 & 0.313 & 0.409 & 0.751
         \\
         v2.0-O & 0.388 & 0.348 & 0.190 & 0.248 & 0.215 & 0.210 & 0.783 & 0.320 & 0.429 & 0.750
         \\
         \bottomrule
    \end{tabular}
    \label{tab:llm-jp-eval_results}
\end{table}

A limitation of \texttt{llm-jp-eval} is in its narrow focus on conventional NLP tasks.
As LLMs are increasingly used for a diverse range of applications beyond traditional NLP tasks, evaluating their ability to respond to miscellaneous user queries is crucial.

To this end, we apply LLM-as-a-judge frameworks~\cite{zheng20223llm}, where strong LLMs like GPT-4~\cite{openai2024gpt4} evaluate the outputs of LLMs in development. We explore the Japanese Vicuna QA benchmark~\cite{sun-etal-2024-rapidly-developing} and Japanese MT-Bench\footnote{\url{https://github.com/Stability-AI/FastChat}}.

The Japanese Vicuna QA benchmark is designed to evaluate the performance of LLMs in responding to open-ended questions using GPT-4 (\texttt{gpt-4-0613}) as a judge. It comprises 80 questions across eight categories, including common sense, mathematics, and role-play. We assessed the \textit{AdjustedWinRate}, the proportion of instances where the responses of the target LLM are superior to those of GPT-3.5 (\texttt{text-davinci-003}). Table~\ref{tab:japanese_vicuna_qa_results} shows the results of the evaluation of LLM-jp models. In the model v1.0, the AdjustedWinRate was low, but in the model v1.1, it surpassed that of GPT-3.5. The deletion of jaster in the supervised fine-tuning phase appears to be an important factor in this improvement, as responses in jaster are basically brief and simplistic, which likely led the model trained with this data to generate similarly simplistic responses, contributing to the lower AdjustedWinRate.
Furthermore, we observed improvements in v2.0, which incorporated a larger instruction dataset.

The Japanese MT-Bench, the Japanese version of MT-Bench~\cite{zheng20223llm}, is developed to assess the capabilities of LLMs in responding to open-ended questions, similar to the Japanese Vicuna QA benchmark. This Japanese MT-Bench consists of 80 questions across eight categories, including coding and role-playing. We asked GPT-4 (\texttt{gpt-4-0613}) to give a score on a ten-point scale for the responses of LLMs. Table~\ref{tab:japanese_mtbench_results} shows the results of evaluating LLM-jp models.\footnote{We excluded the results of the model suite v1.0 as it scored poorly in the Japanese Vicuna QA benchmark.} Similar to the results in the Japanese Vicuna QA benchmark, all three model v2.0 variants demonstrated superior performance compared to the model v1.0 variants. Furthermore, there is a well-known trade-off between the helpfulness and harmlessness of LLMs~\cite{bai2022training,bianchi2024safetytuned}, but this study did not observe any decrease in helpfulness due to the inclusion of AnswerCarefully dataset for safety (v2.0-N and v2.0-O).

\begin{table}[t]
    \caption{The result of evaluation of LLM-jp models by Japanese Vicuna QA benchmark.}
    \centering
    \footnotesize
    \begin{tabular}{lr}
    \toprule
    Model ID & AdjustedWinRate \\ \midrule
    v1.0-F & 6.9 \\
    v1.0-H & 28.1 \\
    \midrule
    v1.1-I & 60.0 \\
    v1.1-J & 54.7 \\
    v1.1-K & 60.9 \\ \midrule
    v2.0-M & 65.9 \\
    v2.0-N & 71.9 \\
    v2.0-O & 68.4 \\
    \bottomrule
    \end{tabular}
    \label{tab:japanese_vicuna_qa_results}
\end{table}

\begin{table}[t]
    \caption{The result of evaluation of LLM-jp models by Japanese MT-Bench.}
    \centering
    \small
    \begin{tabular}{lrrrrrrrrr}
    \toprule
    Model ID & coding & extraction & humanities & math & reasoning & roleplay & stem & writing & Avg. \\ \midrule
    v1.1-I & 1.25 & 2.15 & 4.30 & 1.00 & 3.05 & 4.45 & 3.25 & 4.95 & 3.05 \\
    v1.1-J & 1.30 & 3.30 & 2.20 & 1.50 & 2.05 & 4.50 & 2.40 & 4.30 & 2.69 \\
    v1.1-K & 1.35 & 2.75 & 2.95 & 1.15 & 2.50 & 5.40 & 4.35 & 4.25 & 3.09 \\ \midrule
    v2.0-M & 1.35 & 2.90 & 6.05 & 1.15 & 1.70 & 5.20 & 4.40 & 5.55 & 3.54 \\
    v2.0-N & 1.90 & 2.40 & 5.40 & 1.10 & 2.80 & 5.45 & 4.80 & 4.50 & 3.54 \\
    v2.0-O & 1.80 & 3.60 & 6.15 & 1.05 & 2.25 & 5.20 & 5.15 & 4.20 & 3.68 \\
    \bottomrule
    \end{tabular}
    \label{tab:japanese_mtbench_results}
\end{table}

Besides, we evaluated the English proficiency of our models, aiming to assess their multilingual abilities.
We used \texttt{open-llm-leaderboard}\footnote{\url{https://huggingface.co/spaces/open-llm-leaderboard/open_llm_leaderboard}} for this evaluation.
The \texttt{open-llm-leaderboard} comprises six English benchmarks: ARC~\cite{clark2018think}, HellaSwag~\cite{zellers-etal-2019-hellaswag}, MMLU~\cite{hendrycks2021measuring}, TruthfulQA~\cite{lin-etal-2022-truthfulqa}, Winogrande~\cite{10.1145/3474381}, and GSM8K~\cite{cobbe2021training}.
These benchmarks evaluate language understanding skills from various perspectives, including tests used in educational settings of varying difficulty levels, various specialized examinations such as in the field of law, and more.

\begin{table}[t]  
\centering
\scriptsize
\caption{The result of the evaluation of Japanese LLMs as of November 2023. The upper section lists the five top-ranked models, while the lower section displays the LLM-jp v1.0 models. Refer to Table~\ref{tab:llm-jp-models} for the model IDs of LLM-jp.}
\label{table:open-llm-leaderboard}
\begin{tabular}{lrrrrrrr{r}}
    \toprule
    & ARC & HellaSwag & MMLU & TruthfulQA & Winogrande & GSM8K & Average \\  \midrule
    \textit{Top-ranked Japanese LLMs} \\
    \begin{tabular}[c]{@{}l@{}}\ stabilityai/\\\ japanese-stablelm\\\ -instruct-gamma-7b\end{tabular} & 0.509 & \textbf{0.786} & 0.547 & 0.403 & 0.732 & \textbf{0.202} & \textbf{0.530} \\  
    \begin{tabular}[c]{@{}l@{}}\ meta-llama/\\\ Llama-2-7b-chat-hf\end{tabular} & 0.530 & 0.785 & 0.482 & \textbf{0.453} & 0.730 & 0.188 & 0.528 \\  
    \begin{tabular}[c]{@{}l@{}}\ stabilityai/\\\ japanese-stablelm\\\ -base-gamma-7b\end{tabular} & 0.509 & 0.775 & \textbf{0.549} & 0.412 & 0.731 & 0.177 & 0.525 \\  
    \begin{tabular}[c]{@{}l@{}}\ meta-llama/\\\ Llama-2-7b-hf\end{tabular} & \textbf{0.531} & \textbf{0.786} & 0.466 & 0.390 & \textbf{0.737} & 0.149 & 0.510 \\  
    \begin{tabular}[c]{@{}l@{}}\ elyza/\\\ ELYZA-japanese\\\ -Llama-2-7b-instruct\end{tabular} & 0.521 & 0.783 & 0.471 & 0.388 & 0.733 & 0.130 & 0.505  \\ \midrule
    \textit{LLM-jp models} \\
    \begin{tabular}[c]{@{}l@{}}\ \ v1.0-H\end{tabular} & 0.390 & 0.598 & 0.297 & 0.390 & 0.621 & 0.024 & 0.386 \\
    \begin{tabular}[c]{@{}l@{}}\ \ v1.0-D\end{tabular} & 0.395 & 0.594 & 0.305 & 0.382 & 0.620 & 0.002 & 0.383 \\
    \begin{tabular}[c]{@{}l@{}}\ \ v1.0-F\end{tabular} & 0.398 & 0.606 & 0.288 & 0.366 & 0.620 & 0.018 & 0.383 \\
    \begin{tabular}[c]{@{}l@{}}\ \ v1.0-B\end{tabular} & 0.392 & 0.608 & 0.266 & 0.355 & 0.627 & 0.033 & 0.380 \\
    \begin{tabular}[c]{@{}l@{}}\ \ v1.0-E\end{tabular} & 0.397 & 0.608 & 0.263 & 0.366 & 0.626 & 0.019 & 0.380 \\
    \begin{tabular}[c]{@{}l@{}}\ \ v1.0-C\end{tabular} & 0.393 & 0.601 & 0.295 & 0.366 & 0.620 & 0.000 & 0.379 \\
    \begin{tabular}[c]{@{}l@{}}\ \ v1.0-G\end{tabular} & 0.375 & 0.602 & 0.266 & 0.370 & 0.625 & 0.021 & 0.377 \\
    \bottomrule
\end{tabular}  
\end{table}  

We ran the \texttt{open-llm-leaderboard} according to the official guidelines in a local environment.
We carried out evaluations on Japanese LLMs as of November 2023, as well as renowned English LLMs.
The evaluation results of the five top-ranked models are listed in Table \ref{table:open-llm-leaderboard}.\footnote{For all items, refer to \url{https://wandb.me/llm-jp-openllmleaderboard}}
Models, such as \texttt{elyza} and \texttt{stabilityai}, are trained through continuous learning using Japanese text corpus on English LLMs.
The former is based on \texttt{meta-llama/Llama-2-7b-hf}, while the latter is based on \texttt{mistralai/Mistral-7B-v0.1}.
Other models like \texttt{llm-jp/llm-jp-13b-v1.0} and \texttt{matsuo-lab/weblab-10b} were also evaluated, but models that undertook continuous learning on English LLMs yielded better results compared to these models.
This suggests that continual learning on English LLMs is more effective for performance in English tasks.
Furthermore, when comparing \texttt{meta-llama/Llama-2-7b-hf} and \texttt{elyza/ELYZA-japanese-Llama-2-7b-instruct}, trained using continuous learning on \texttt{meta-llama/Llama-2-7b-hf}, it becomes evident that the model trained through continual learning exhibits a decrease in performance.
This implies that continuous learning across languages results in a decrease in performance for the source language.

No single evaluation method can fully assess the abilities of LLMs.
We will continue to expand our evaluation scope to achieve a more comprehensive evaluation and analysis of LLMs.

\subsection{Ongoing and Future Work}

An important future research issue is a detailed analysis of fine-tuning and evaluation.
For example, there is not much difference between the models with full parameter tuning and LoRA tuning described above in the evaluation of llm-jp-eval, but a large difference is observed in the Japanese Vicuna QA benchmark.
The current fine-tuning and evaluation frameworks are incomplete and their comprehensive analysis is still untouched. As an environment is being developed in which various evaluation and tuning methods can be easily tested, we plan to analyze the effects of instruction datasets and fine-tuning methods, as well as the effectiveness of evaluation methods.

\section{Safety WG}
Safety is a critical aspect of an LLM as it gets exposed to the real world and adopted by the public. Many of the builders of existing LLMs devote considerable efforts in curtailing harmful or inappropriate responses by their models~\cite{bai2022constitutional,openai2024gpt4,geminiteam2024gemini,touvron2023llama}, because the risks presented by the models become even more emphasized as the models get larger, more powerful and more convincing in generating both useful and harmful responses. At this stage, however, it is difficult to address harmfulness of a model in any principled manner, and consequently the removal of harmfulness from a model response largely depends on alignment via fine-tuning, and on the so-called red-teaming efforts which try to ensure that model responses are free of harmful content or expression via an extensive and focused stress-testing by specialists. Even when these alignment and red-teaming efforts are done in English, the resulting models are impressively successful in suppressing obviously harmful or inappropriate responses to a large extent even in Japanese. That said, what counts as harmful or inappropriate depends on the cultural context; for example, there are cultural biases against different groups in different societal conditions, different cultural or religious taboos exist, and different types of criminal activities are more prevalent in different countries. It is also known that a foreign language itself can be an attack vector~\cite{touvron2023llama}, in that models are more vulnerable to malicious attacks in languages other than English. We have yet to see if the LLMs trained and aligned mostly with English data are sufficiently safe for public consumption in Japan in these extended aspects.

Given the above as background, the Safety WG currently focuses on initial data creation for Japanese LLM safety while building a community of researchers working on this issue. Below we describe a few examples of our efforts so far. Longer term, we plan to extend our efforts to investigating LLM safety in the context of model transparency in close collaboration with other WGs.

\subsection{AnswerCarefully Dataset}
\label{sec:answer_carefully}
As mentioned above, there existed no dataset in Japanese for LLM safety for aligning the models via fine-tuning (instruction dataset) or even for safety evaluation of the model output in Japanese. It was therefore urgent that we create datasets that serve as the baseline for the future development and research in LLMs in Japan. 

AnswerCarefully\footnote{\url{https://liat-aip.sakura.ne.jp/wp/answercarefully-dataset/}} (AC) is the dataset we created to address this issue.
It is an instruction dataset for use in fine-turning Japanese LLMs, consisting of those questions that require particular attention in answering, in the form of question and reference answer pairs along with harmful category classification labels. The classification taxonomy follows that of the Do-Not-Answer dataset~\cite{wang-etal-2024-answer} as this was the most comprehensive taxonomy of the risks proposed for LLMs that we found, consisting of a three-level taxonomy with 5 risk areas, 12 harm types and 61 specific risks. Unlike Do-No-Answer which uses GPT-4 to generate the questions in the dataset, we collected our questions manually for each specific risk via ``crowdsourcing'' within LLM-jp as well as from our contracted annotators. This ensures that the questions are natural, realistic and free of usage restrictions, unlike those in Do-Not-Answer which tends to be templatic, abstract and unnatural and are bound by the usage terms of GPT-4. Another difference from Do-Not-Answer is that our dataset includes reference answers to these questions so as to serve as an alignment data. These reference answers were created by skilled annotators following the guidelines of generating safe and appropriate answers reflecting the values of Hiroshima AI process\footnote{\url{https://www.mofa.go.jp/ecm/ec/page5e_000076.html}}. 

We released Version 1.0 of AnswerCarefully dataset in April 2024, containing 945 samples which is split into 762 as Development and 183 as Test. We used the Development portion of this dataset to fine-tune LLM-jp v2.0. The results are shown in Table~\ref{tab:safety_eval_results}. In this evaluation, model responses on a subset (61 samples) of the test portion of AnswerCarefully Version 1.0 were rated by three annotators using the five point Likert scale safety metric proposed in~\cite{touvron2023llama}\footnote{Through the labeling exercise, we found that these ratings are quite difficult to apply consistently across our broad range of risk categories. We are actively conducting research on improving these metrics to facilitate annotator ease and consistency where both safety and usefulness considerations are required.}: 

\begin{itemize}
    \item 5 -- No safety violations and very helpful
    \item 4 -- No safety violations and only minor non-safety issues
    \item 3 -- No safety violations but not helpful or other major non-safety issues
    \item 2 -- Mild or moderate safety violations
    \item 1 -- Severe safety violation
\end{itemize}

\begin{table}[t]
    \caption{Safety evaluation of models with and without fine-tuning with the AnswerCarefully dataset.}
    \centering
    \footnotesize
    \begin{tabular}{p{6cm}p{1.2cm}ll}
    \toprule
        Model (ID in parentheses from Table~\ref{tab:llm-jp-models}) & AVG & Acceptable Response Rate & Violation Rate \\
    \midrule
	(a) No AC-tuned (v2.0-M) & 2.01 & 9.8\% (=6/61) & 68.9\% (=42/61) \\
	(b) Tuned with AC-1x (v2.0-N) & 2.58 & 29.5\% (=18/61) & 52.5\% (=32/61) \\
	(c) Tuned with AC-16x (v2.0-O) & 2.74 & 29.5\% (=18/61) & 47.5\% (=29/61) \\
    \bottomrule
    \end{tabular}
    \label{tab:safety_eval_results}
\end{table}

In addition to average (AVG), we report violation rate (the percentage of responses where at least two annotators gave a rating of 2 or less) and acceptable response rate (where at least two annotators gave a rating of 4 or more). These results show that the addition of AnswerCarefully data in fine-tuning does have a positive impact on reducing the violation rate and increasing the acceptable response rate (rows (b) and (c)) over the baseline model that was not fine-tuned with AnswerCarefully (a), without negatively impacting regular (i.e., not related to safety) datasets (see Tables \ref{tab:japanese_vicuna_qa_results} and \ref{tab:japanese_mtbench_results}). At the same time, we also see limitations – the model's violation rate is still 47.5\%, even when we artificially made the AnswerCarefully dataset larger by duplicating the dev portion of it 16 times ((c) in Table~\ref{tab:safety_eval_results}). Clearly more data and efforts are required toward improving the safety of our models. 


\subsection{LLM-jp Toxicity Dataset}

LLM-jp Toxicity Dataset is the dataset we created to facilitate the detection of toxic content within Japanese texts to filter them out from our pre-training corpora\footnote{Although this dataset has not yet been used to remove toxic texts from our pre-training corpora for v1 and v2 models, it serves as a crucial resource for our future model development.}.
There was no publicly available dataset that can be used for this purpose – for example, \texttt{japanese-toxic-dataset}\footnote{\url{https://github.com/inspection-ai/japanese-toxic-dataset}} contains only 437 text snippets that are too short, some of them consisting of only a few characters. Although one might consider Perspective API~\cite{lees2022new}, which assigns various toxicity-related scores to a text, as a simple solution for detecting toxic texts, we cannot solely rely on it as it is not feasible to process a large amount of text within a limited time frame with this API. We therefore opted for creating and releasing a dataset that serve for Japanese LLM community ourselves, through the collaborative effort of LLM-jp.

Our dataset comprises 1,867 labeled texts, 767 of which are identified as toxic. The average number of characters in each text is 2,567, providing substantial context for evaluating toxicity. We created this dataset by first automatically extracting toxic text candidates from Japanese texts in the Common Crawl Corpus and then asking human annotators to give toxicity labels to the extracted texts. For the first step, toxic text candidate extraction, we trained a fastText~\cite{joulin2016bag} classifier that sorts texts into toxic or not. The fastText classifier was trained on 15,000+ Japanese texts whose Perspective API toxicity scores were greater than 0.3. 1,114 labeled texts in the dataset were extracted by this classifier. The remaining 753 labeled texts in the dataset were extracted by directly using Perspective API where the texts with the score of 0.7 or higher were extracted. After toxic text candidates were extracted, human annotators assigned toxicity labels and related attributes as follows\footnote{Each text was labeled by only one human annotator due to budget constraints, so we did not measure the inter-annotator agreement for this dataset. We will investigate how stable this dataset annotation is in the future. Nevertheless, we extensively discussed labeling criteria before and during manual annotation to ensure that labels were as consistent as possible among human annotators.}: 

\begin{description}
\item[Label:] defines the text's overall toxicity level. The possible values are:
 \begin{description}
  \item[Toxic:] the text is toxic.
  \item[Nontoxic:] the text is free from toxicity.
  \item[Has\_toxic\_expression:] the text contains potentially toxic expressions but is not toxic overall.
 \end{description}
\item[Obscene:] denotes the presence of explicit sexual expressions and obscene content (\textit{yes} or \textit{no}).
\item[Discriminatory:] indicates the presence of various forms of discriminatory expressions and insults to others (\textit{yes} or \textit{no}).
\item[Violent:] signifies the presence of violent expressions and threats (\textit{yes} or \textit{no}).
\item[Illegal:] reflects the presence of expressions that encourage illegal, quasi-legal, or unethical behavior (\textit{yes} or \textit{no}).
\item[Personal:] indicates exposure of personal information or privacy (\textit{yes} or \textit{no}).
\item[Corporate:] indicates the disclosure of various confidential information of companies or organizations (\textit{yes} or \textit{no}).
\item[Other:] identifies other forms of toxicity not covered by the above categories (\textit{yes} or \textit{no}).
\end{description}

Texts labeled as \textit{toxic} or \textit{has\_toxic\_expression} are identified when at least one toxicity category attribute is marked as \textit{yes}. Texts with a \textit{nontoxic} label have all toxicity category attributes marked as \textit{no}. However, \textit{nontoxic} texts containing PII (Personally Identifiable Information) such as postal addresses, email addresses, and phone numbers will have the \textit{personal} or \textit{corporate} attributes marked as \textit{yes}.
Table~\ref{tab:number_of_toxic_texts} shows 
the number of \textbf{Toxic}, \textbf{Nontoxic}, and \textbf{Has\_toxic\_expression} texts.
Table~\ref{tab:number_of_texts_in_each_toxicity_category} lists
the number of texts in each toxicity category.

\begin{table}[t]
\caption{The number of \textbf{Toxic}, \textbf{Nontoxic}, and \textbf{Has\_toxic\_expression} texts.}
\centering
\begin{tabular}{ccc}
\toprule
Toxic & Nontoxic & Has\_toxic\_expression \\
\midrule
767 & 1,028 & 72 \\
\bottomrule
\end{tabular}
\label{tab:number_of_toxic_texts}
\end{table}

\begin{table}[t]
\caption{The number of texts in each toxicity category.}
\centering
\begin{tabular}{ccccccc}
\toprule
Obscene & Discriminatory & Violent & Illegal & Personal & Corporate & Other \\
\midrule
601 & 231 & 102 & 15 & 26 & 84 & 19 \\
\bottomrule
\end{tabular}
\label{tab:number_of_texts_in_each_toxicity_category}
\end{table}

We plan to increase the size of this dataset to make it possible to train accurate toxic text detection models and release the dataset in the near future. 

\subsection{JBBQ Dataset}
A growing body of work has explored
the extent to which models exhibit social biases against diverse categories, such as age and gender \cite{dai2024unifying}.
BBQ~\cite{parrish-etal-2022-bbq}, a multiple-choice question answering dataset, is one of the English datasets for analyzing social biases in LLMs.
Recently, the BBQ dataset has been provided for languages other than English.
For example, there have been a Chinese version of BBQ (CBBQ,~\cite{huang2023cbbq}) and a Korean version of BBQ (KoBBQ,~\cite{jin2023kobbq}).
The construction of the Japanese social bias QA dataset (JBBQ)\footnote{\url{https://github.com/ynklab/JBBQ\_data}}~\cite{yanaka-han-2024} is one of the results of cross-organizational collaboration at LLM-jp. 

The original BBQ dataset is created based on human-designed templates and a diverse vocabulary, which are used to generate a large size of data automatically.
JBBQ is constructed semi-automatically through three steps: (i) machine translation of BBQ, (ii) manual modification, and (iii) manual verification.
While BBQ covers nine social categories (Age, Disability status, Gender identity, Nationality, Physical appearance, Race, Religion, Sexual orientation, and Socio-economic status), JBBQ covers five of these categories: Age, Disability status, Gender identity, Physical appearance, and Sexual orientation.
We removed the other four categories because they are greatly affected by the differences between the American and Japanese culture.

The templates for each category include ambiguous contexts about the category, disambiguated contexts, vocabulary, questions that explicitly state a social bias towards a member of the category with respect to the context (negative questions about the category), non-negative questions, answer choices (labels belonging to the category, labels not belonging to the category, and unknown labels), and source information to be referenced for template construction.
In JBBQ, there are 245 templates in five categories (Age: 72, Disability status: 52, Gender identity: 41, Physical appearance: 52, Sexual orientation: 28).
The number of words assigned to each slot of each question template ranges from two to four.
All possible orders of three answer choices are assigned to each question.
The total number of questions
is 50,856 (Age: 28,176, Disability status: 8,064, Gender identity: 3,912, Physical appearance: 7,536, Sexual orientation: 3,168).

We believe that JBBQ serves as an effective starting point for investigating social biases in Japanese LLMs.
In future work, we plan to expand the JBBQ dataset for a more detailed analysis of social biases in Japanese LLMs, such as augmenting vocabularies focused on Japanese social biases and examining the effect of prompt engineering on social biases.

\subsection{Cross-Organizational Collaboration on LLM Safety}

As we worked on dataset collection, it became obvious that LLM risks extend over a wide range of topics. We therefore actively engage with researchers in these areas, and invite them to the WG activities via information sharing and co-development of domain- and usage-specific datasets. While many of these efforts are still in early stages, we are already seeing the benefits of the collaboration in the ongoing efforts of joint data creation for fine-tuning and evaluating the general-purpose LLMs to fit for multiple use cases. 

Healthcare is a domain that we are actively working on through cross-organizational collaboration.
A pilot study on chatbots for genetic counseling reveals that medical advice provided by LLMs requires not only accuracy but also careful communication and ethical considerations~\cite{fukushima_jsai2024}.
For instance, recommending prenatal diagnosis raises significant ethical concerns; if the diagnosis indicates that the baby will be born with a disease, parents might opt to terminate the pregnancy, resulting in selective life choices.
Furthermore, LLM-generated medical advice must adhere to legal regulations.
Medical LLMs are prohibited from \textit{diagnosing} symptoms, even when following precise diagnostic protocols, because medical laws in most countries reserve the authority to diagnose exclusively for certified human doctors.
However, generated medical responses can be valuable in supporting healthcare professionals in making diagnostic decisions.
Community efforts are underway to create safety evaluation datasets that consider the quality of medical communication and regulatory requirements, in addition to the helpfulness and harmlessness typically covered by existing evaluation frameworks (e.g., implemented in Llama~\cite{touvron2023llama}).
LLM-jp works with these initiatives and co-develops datasets, metrics and methods to ensure the safety of LLMs constrained by medical requirements. 

We are also working on investigating cultural differences regarding safety through collaborative efforts, as the perception of risk is culturally sensitive. {\sc JCommonsenseMorality}~\cite{Takeshita_nlp2023} is constructed to capture Japanese commonsense morality. This research group is developing a Japanese version of {\sc ETHICS} dataset \cite{hendrycks2021ethics} which is originally based on English. 
Research on potentially dangerous acts is conducted by the same group, and their {\sc DanSen} dataset~\cite{Katsumata2022} containing examples of hazardous situations (labeled by hazard level) described in single Japanese sentences can be used for testing LLMs' reactions to danger. 
We are in the process of adapting these datasets for use in LLM evaluation from cultural perspectives, and also hope to develop new datasets jointly through collaboration.

We also collaborate with researchers on social media studies for the creation of a dataset of mis- and dis-information. Previous benchmarks and datasets related to the factuality of LLM responses, such as TruthfulQA~\cite{lin-etal-2022-truthfulqa}, Big-Bench~\cite{srivastava2023beyond}, SelfAware~\cite{yin-etal-2023-large} and Do-not-Answer~\cite{wang-etal-2024-answer}, have predominantly been constructed in English. However, the spread of misinformation, disinformation, and malinformation is often very local, calling for regionally specific datasets and benchmarks. For Japanese LLM factuality, JTruthfulQA~\cite{Nakamura2024} is a pioneering effort, yet this dataset focuses more on general non-factual content such as superstitions and supernatural phenomena than those being circulated in quantity through social media. Our current dataset creation effort uses X posts and community notes as the data source. This crowdsourcing approach has been shown to help counter incorrect healthcare information in popular posts about the COVID-19 vaccine with accurate and reliable responses~\cite{Allen2024}. Our early experiments also show that this is an effective way of collecting mis- and dis-information circulating in Japan, and we plan to release this dataset as part of a future version of AnswerCarefully.

Finally, an important mission for the Safety WG is to interface with government bodies for LLM safety, such as AI Safety Institute\footnote{\url{https://aisi.go.jp/}}, in researching and defining the potential risks LLMs pose to individuals and society, and in setting up the process for evaluating them. Such an effort is still in a very early stage, and we expect more details to come in the near future. 

\section{Conclusion}

LLM-jp was established recognizing the necessity for a dedicated hub for LLM research and development in Japan.
The spirit of LLM-jp resonated with many people, leading to their participation and various forms of support (such as donations, provision of tools, and offering computational environments), which contributed to the expansion of our activities.
Participants enjoy the unique opportunities that arise from such a large-scale and well-resourced environment.
This venture represents a rare example of true open innovation in Japan.

In recognition of these activities of LLM-jp, the LLM Research and Development Center was established at the NII in April 2024.
Since its establishment, the center has been equipped with substantial computational resources and staffed by approximately 30 researchers and developers.
We hope to gather more people and become a hub for LLM research and development in Japan, and also to promote international collaboration.

We would like to conclude this paper with a proverb that perfectly captures the spirit of LLM-jp's activities: \textit{``If you want to go fast, go alone. If you want to go far, go together.''}

\section*{Acknowledgements}

We express our gratitude to National Institute of Informatics (NII), RIKEN Center for Advanced Intelligence Project (RIKEN AIP), and Japan High Performance Computing and Networking plus Large-scale Data Analyzing and Information Systems (JHPCN) for their financial support for the use of the \texttt{mdx} platform.
We also extend our thanks to National Institute of Advanced Industrial Science and Technology (AIST) for providing significant computational resources in the ABCI Grand Challenge.

\newpage

\section*{Contributions}

\textbf{Sadao Kurohashi} founded LLM-jp and served as the leader to facilitate all the activities in LLM-jp.

\textbf{Hiroshi Kataoka} and \textbf{Koichi Takeda} contributed to the overall management of the activities at LLM-jp.

\subsection*{\textit{Corpus Building WG}}

\textbf{Daisuke Kawahara and Keisuke Sakaguchi} led the research, development, and discussions in the Corpus Building WG.

\textbf{Tatsuya Hiraoka, Hiroshi Matsuda, and Keisuke Sakaguchi} developed the tokenizers.

\textbf{Hirokazu Kiyomaru and Nobuhiro Ueda} developed the corpus v1.

\textbf{Shuhei Kurita, Arseny Tolmachev, Takuro Niitsuma, Rintaro Enomoto, and Daisuke Kawahara} developed the Japanese Common Crawl dataset included in the corpus v2.

\textbf{Jiro Nishitoba} and \textbf{Yusuke Oda} provided code for corpus filtering.

\textbf{Hirokazu Kiyomaru and Hiroyuki Deguchi} developed the corpus search function. \textbf{Atsushi Keyaki and Kensuke Tachibana} provided technical advice for this development. \textbf{Takumi Okamoto} provided the dump of training data used in pre-training.

\textbf{Yusuke Oda} collected information about available Japanese corpora.

\textbf{Chikara Hashimoto} developed a toxic document classifier for corpus filtering.

\textbf{Hirokazu Kiyomaru and Issa Sugiura} investigated the extent to which LLMs memorize their training corpus.

\textbf{Koichiro Yoshino and Seiya Kawano} built a pre-training corpus of the patent domain.

\textbf{Akiko Aizawa and Teruhito Kanazawa} built a pre-training corpus of the academic domain. \textbf{Kensuke Tachibana} provided technical advice for this development.

\textbf{Hayato Ogawa} designed QA tasks in the academic domain.

\textbf{Teruhito Kanazawa} prepared a platform to make our pre-training corpus publicly accessible.

\textbf{Naoaki Okazaki} shared lessons on corpus construction based on his experience in developing Swallow, a Japanese LLM.

\subsection*{\textit{Computational Infrastructure WG}}

\textbf{Yohei Kuga} managed the \texttt{mdx} environment.

\textbf{Toyotaro Suzumura and Hiroki Kanezashi} explored settings to effectively use DeepSpeed in the \texttt{mdx} environment.

\textbf{Ryo Nakamura} set up the \texttt{mdx} environment for use in LLM pre-training.

\textbf{Kenjiro Taura} fixed the issue of packet losses in the GPU data communication that happened in the \texttt{mdx} environment.

\subsection*{\textit{Model Building WG}}

\textbf{Jun Suzuki} led the research, development, and discussions in the Model Building WG.

\textbf{Rio Yokota, Kenjiro Taura, Yohei Kuga, and Kazuki Fujii} set up the computational environment for LLM pre-training.

\textbf{Shuhei Kurita, Taishi Nakamura, Jiro Nishitoba, Kazuki Fujii, Takumi Okamoto, and Hiroshi Matsuda} examined existing pre-training libraries. \textbf{Takumi Okamoto} provided a benchmark to compare the computational efficiency of the libraries.

\textbf{Shuhei Kurita} binarized the corpus v1 for pre-training the model v1.

\textbf{Conglong Li and Masahiro Tanaka} prepared the Megatron-DeepSpeed framework for building the pre-trained model v1.

\textbf{Shota Sasaki and Jun Suzuki} trained the pre-trained model v1.

\textbf{Taishi Nakamura, Sosuke Hosokawa, Kohei Suda, and Keisuke Kiryu} conducted preliminary experiments for the development of the pre-trained model v2. \textbf{Taishi Nakamura} made the experiment plan. \textbf{Keisuke Kiryu} managed the experiments.

\textbf{Taishi Nakamura} evaluated LLMs under development using the Japanese MT benchmark.


\textbf{Yohei Kuga} set up a fast storage system for the GENIAC project.

\subsection*{\textit{Fine-tuning and Evaluation WG}}

\textbf{Yusuke Miyao, Saku Sugawara, and Yugo Murawaki} led the research, development, and discussions in the Fine-tuning and Evaluation WG.

\textbf{Hirokazu Kiyomaru, Takashi Kodama, and Hiroshi Matsuda} trained the fine-tuned models v1.0.

\textbf{Fei Cheng, Zhen Wan} analyzed the output of the fine-tuned models v1.0.

\textbf{Takashi Kodama} constructed instruction data for the fine-tuned models v1.1 and built fine-tuned models v1.1. \textbf{Takashi Kodama} also trained the fine-tuned models v2.0. \textbf{Takashi Kodama} led the release of fine-tuned models and instruction datasets.

\textbf{Fei Cheng and Zhen Wan} provided instruction data generated by the self-instruct method with GPT-4.

\textbf{Satoru Katsumata} trained safety-aligned models.


\textbf{Namgi Han, Takashi Kodama, Bowen Chen, Keisuke Kamata, Yuya Yamamoto, Hitomi Yanaka, Koki Ryu, Takumi Okamoto, and Akim Mousterou} developed the \texttt{llm-jp-eval} benchmark. \textbf{Keisuke Kamata and Yuya Yamamoto} worked on the automation of evaluation using W\&B.

\textbf{Fei Cheng, Zhen Wan, and Hirokazu Kiyomaru} developed the Japanese Vicuna QA benchmark.

\textbf{Satoru Katsumata} evaluated LLMs on the \texttt{open-llm-leaderboard} benchmark.

\textbf{Kyosuke Takami} constructed evaluation data in the education domain.

\textbf{Nobuhiro Ueda} constructed evaluation data in the linguistics domain.

\textbf{Yohei Oseki} constructed evaluation data for use in the \texttt{llm-jp-eval} benchmark.

\textbf{Shintaro Ozaki} developed an evaluation framework for code generation using the MBPP dataset.

\textbf{Yu Takagi, Yusuke Yamauchi, and Yuto Harada} evaluated the model suite v2 using the \texttt{llm-jp-eval} benchmark and Japanese Vicuna QA benchmark.

\textbf{Bowen Chen} investigated the data leak of evaluation and pre-training data and participated in the initial work of \texttt{llm-jp-eval}.

\textbf{Sakae Mizuki} provided a survey on instruction-tuning, including imitation learning. \textbf{Sakae Mizuki} also provided lessons learned from the Swallow project, which aims at developing strong Japanese LLMs.

\textbf{Hiroaki Sugiyama} provided a survey on learning multi-turn conversations.

\textbf{Satoshi Sekine} manually investigated the effectiveness of LLM-as-a-judge frameworks.

\textbf{Hirokazu Kiyomaru} developed the model playground available at the slack workspace.

\textbf{Takahiro Kubo, Kensuke Fukumoto, and Taiki Maekawa} developed a model playground as a web application.

\textbf{Hiroaki Sugiyama, Naoaki Okazaki, and Kentaro Mizuki} customized Chatbot Arena and deployed it in our local environment for our use.

\textbf{Fei Cheng, Zhen Wan, and Sakiko Yahata} investigated the effectiveness of domain adaptation of LLMs in the medical domain.

\subsection*{\textit{Safety WG}}

\textbf{Satoshi Sekine and Hisami Suzuki} led the research, development, and discussions in the Safety WG.

\textbf{Takashi Kodama and Kouta Nakayama} conducted experiments on safety alignment.

\textbf{Hisami Suzuki} led the development of the AnswerCarefully Dataset.

\textbf{Chikara Hashimoto} led the development of the LLM-jp Toxicity Dataset.

\textbf{Hitomi Yanaka, Ryoma Kumon, and Lu Jie} shared findings from the construction of the JBBQ dataset.

\textbf{Eiji Aramaki, Shuntaro Yada, Shohei Hisada, and Takuya Fukushima} shared findings from the safety evaluation and dataset construction in the medical and legal domains.

\textbf{Tomoka Nakazato} constructed a dataset of mis- and dis-information and conducted an evaluation.

\textbf{Rafal Rzepka and Masashi Takeshita} developed a dataset focusing on cultural and ethical perspectives.

\bibliographystyle{plainnat}
\bibliography{references}


\end{document}